\documentclass[runningheads]{llncs}

\usepackage{accv}

\usepackage{accvabbrv}

\usepackage{graphicx}
\usepackage{booktabs}
\usepackage{kotex}
\usepackage[ruled]{algorithm2e}
\usepackage{mathtools}
\usepackage{makecell}
\usepackage{multirow}
\usepackage{graphicx}
\usepackage{subcaption}
\usepackage{mwe} 
\usepackage{wrapfig}
\usepackage[accsupp]{axessibility}  

\usepackage{hyperref}

\usepackage{orcidlink}

\begin{document}

\makeatletter
\newcommand{\printfnsymbol}[1]{%
  \textsuperscript{\@fnsymbol{#1}}%
}
\makeatother

\title{RNA: Video Editing with ROI-based Neural Atlas} 

\titlerunning{RNA}

\author{Jaekyeong Lee\thanks{equal contribution} \and
Geonung Kim\printfnsymbol{1}  \and
Sunghyun Cho} 

\authorrunning{J.~Lee et al.}

\institute{POSTECH\\
\email{\{jaekyeong,k2woong92,s.cho\}@postech.ac.kr}\\
\url{https://jaekyeongg.github.io/RNA}
}


\maketitle
\begin{center}
    \centering
    \captionsetup{type=figure}
     \vspace{-5mm}
    \includegraphics[width=0.99\textwidth]{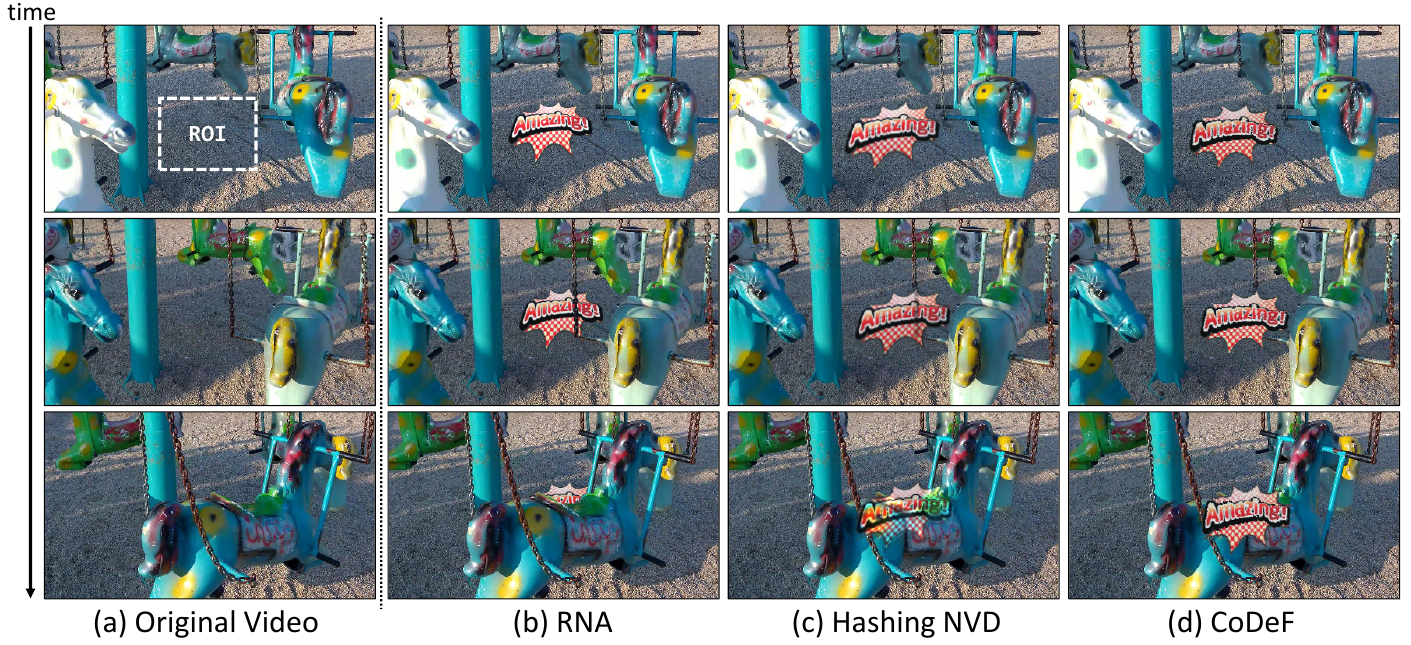}
    \vspace{-2mm}
    \captionof{figure}{
    (b) Our video editing achieves natural editing outcomes, successfully considering the occlusions from the thin chain (2nd row) and the toy horse of the carousel (3rd row). In contrast, (c) Hashing NVD~\cite{hashingnvd} results in ghosting artifacts, and (d) CoDeF~\cite{codef} neglects the occlusion from moving objects, failing to produce natural editing results.
    }
    \vspace{-2mm}
   \label{fig:teaser}
\end{center}

\def\MethodName{RNA} 

\newcommand{\Eq}[1]  {Eq.\ (#1)}
\newcommand{\Eqs}[1] {Eqs.\ (#1)}
\newcommand{\Fig}[1] {Fig.\ #1}
\newcommand{\Figs}[1]{Figs.\ #1}
\newcommand{\Tbl}[1]  {Tab.\ #1}
\newcommand{\Tbls}[1] {Tabs.\ #1}
\newcommand{\Sec}[1] {Sec.\ #1}
\newcommand{\SSec}[1] {Sec.\ #1}
\newcommand{\Secs}[1] {Secs.\ #1}
\newcommand{\Alg}[1] {Alg.\ #1}

\newcommand{\setone}[1] {\left\{ #1 \right\}} 
\newcommand{\settwo}[2] {\left\{ #1 \mid #2 \right\}} 

\newcommand{\kkw}[1]{{\textcolor[rgb]{0.0,0.6,0.4}{[KGU: #1]}}}
\newcommand{\ljk}[1]{{\textcolor[rgb]{0.4,0.6,0.0}{LJK: #1}}}
\newcommand{\sunghyun}[1]{{\textcolor[rgb]{0.6,0.0,0.6}{sunghyun: #1}}}
\newcommand{\eh}[1]{{\textcolor[rgb]{0.0,0.6,0.6}{eh: #1}}}

\newcommand{\bb}[1]{\textbf{\textit{#1}}}
\newcommand{\xx}{\textcolor{red}{XX}}
\newcommand{\fix}[1]{\textcolor{red}{#1}}
\newcommand{\rt}[1]{\textcolor{red}{#1}}
\newcommand{\bt}[1]{\textcolor{blue}{#1}}

\newcommand\Myperm[2][^n]{\prescript{#1\mkern-2.5mu}{}P_{#2}}
\newcommand\Mycomb[2][^n]{\prescript{#1\mkern-0.5mu}{}C_{#2}}






\begin{abstract}

With the recent growth of video-based Social Network Service (SNS) platforms, the demand for video editing among common users has increased. However, video editing can be challenging due to the temporally-varying factors such as camera movement and moving objects. While modern atlas-based video editing methods have addressed these issues, they often fail to edit videos including complex motion or multiple moving objects, and demand excessive computational cost, even for very simple edits. In this paper, we propose a novel region-of-interest (ROI)-based video editing framework: ROI-based Neural Atlas (RNA). Unlike prior work, RNA allows users to specify editing regions, simplifying the editing process by removing the need for foreground separation and atlas modeling for foreground objects. However, this simplification presents a unique challenge: acquiring a mask that effectively handles occlusions in the edited area caused by moving objects, without relying on an additional segmentation model. To tackle this, we propose a novel mask refinement approach designed for this specific challenge. Moreover, we introduce a soft neural atlas model for video reconstruction to ensure high-quality editing results. Extensive experiments show that RNA offers a more practical and efficient editing solution, applicable to a wider range of videos with superior quality compared to prior methods.


\end{abstract}


\section{Introduction} 
\label{sec:intro}

With the recent growth of video-based Social Network Service (SNS) platforms such as YouTube Shorts, there has been an explosive increase in the demand for video editing among common users. 
However, video editing is a challenging task since videos have temporally-varying characteristics caused by various factors such as camera movement and moving objects, and editing such videos must involve addressing these tricky factors in a temporally-consistent way.

For temporally-consistent video editing, atlas-based video editing methods such as Layered Neural Atlas (LNA)~\cite{lna}, Hashing Neural Video Decomposition (Hashing NVD)~\cite{hashingnvd}, and Content Deformation Fields (CoDeF)~\cite{codef} have been proposed.
These methods operate under an assumption, in which each video frame consists of sprite layers representing the background and foreground objects, and the appearance of the sprites are constant, while their positions and shapes change by their motions, which can be represented, e.g., by optical flow \cite{opticalflow}.
Based on this assumption, the typical procedure of atlas-based approaches is as follows.
First, they separate an input video into foreground objects and the background using instance segmentation, then estimate the atlases and motions of the separated regions.
Finally, a user edits these atlases, and then an edited video is reconstructed from the edited atlases and the motion data.



Unfortunately, the atlas-based video editing approaches have inherent limitations caused by their strategy that explicitly models each foreground object independently.
Firstly, accurate instance segmentation and atlas estimation of multiple foreground objects are challenging especially when foreground objects have complex motions.
Failures in segmentation and atlas estimation result in error-prone editing outcomes, e.g., ghosting artifacts or failures of handling occlusions in \cref{fig:teaser} (c) and (d).
Secondly, for the segmentation of foreground objects, the atlas-based video editing approaches require users to specify all the foreground objects including those that the users do not intend to edit.
This requirement significantly diminishes the user-friendliness of the video editing interface, particularly when the input video contains multiple foreground objects.
Lastly, in practical scenarios, users frequently aim to modify only a specific region of a video, such as introducing a new object or altering the texture of an existing one. 
Despite this, prior atlas-based video editing methods, such as LNA~\cite{lna} and Hashing NVD~\cite{hashingnvd}, necessitate the complete reconstruction of the entire video, leading to substantial computational resource requirements. 
In particular, they demand memory space and computation time that scale with the number of objects. This results in computation times spanning several hours for videos containing multiple objects.


This paper proposes a novel region-of-interest (ROI)-based video editing framework: ROI-based Neural Atlas (RNA).
Unlike previous methods, RNA allows users to specify a region where editing will occur and then optimizes a single atlas exclusively for the specified region. After the optimization, editing is performed directly on the atlas. Finally, RNA reconstructs an edited video using the edited atlas for the ROI, and the original pixel values for the non-edited regions.
Our ROI-based approach enjoys a couple of advantages over prior methods. Firstly, our approach does not need foreground separation performed by segmentation models, which are often unreliable and cumbersome for users. Additionally, by focusing solely on an ROI atlas, it avoids the necessity of atlas estimation for all moving foreground objects, thus enabling editing of videos with complex motions without incurring ghosting artifacts, and maintaining constant computational resources regardless of the number of foreground objects.

Eliminating the foreground separation process introduces a unique challenge that sets it apart from previous methods: acquiring a mask that effectively addresses occlusions in the edited area caused by moving objects, without depending on an additional segmentation model.
To this end, we propose a novel approach for mask refinement tailored to the specific challenge.
Specifically, after estimating the atlas and mask similarly to previous methods~\cite{lna,hashingnvd}, we further refine the imperfect mask using a novel self-supervised method.
Additionally, we introduce a novel soft neural atlas model, which employs a soft mask for handling boundaries between occluding objects and an edited atlas for high-quality video reconstruction.

Our main contributions can be summarized as follows.
\vspace{-2mm}
\begin{itemize}
\item We first propose a novel ROI-based video editing framework that utilizes a single atlas for editing regions, which removes requirement for the foreground separation process and atlas modeling for all foreground objects.
\item To estimate an accurate mask without foreground separation, we propose a novel mask refinement method.
\item We also introduce a novel soft neural atlas model for more natural-looking video reconstruction.
\item Extensive experiments demonstrate the efficiency and effectiveness of our video editing framework, underscoring its practicality and versatility, compared to previous state-of-the-art methods.
\end{itemize}

\section{Related Work} 
\label{sec:relatedwork}

\paragraph{Atlas-based Video Editing}
Video editing via 2D atlas images was first introduced by Unwrap Mosaics~\cite{unwrap}, which decomposes a video into a series of 2D texture maps termed ``unwrap mosaics'', and establishes a mapping from these mosaics to video frames. 
Editing is then directly applied to the unwrap mosaics. 
To enhance this approach, which involves complex optimization, na\"{i}ve layering with binary segmentation masks, and limited adaptability for non-rigid objects, LNA~\cite{lna} introduces an end-to-end self-supervised method that reconstructs videos using neural atlases with alpha blending. 
To expedite the optimization, Hashing NVD\cite{hashingnvd} and CoDeF~\cite{codef} adopt hash encoding~\cite{instantngp} to represent input videos.
To facilitate more advanced editing, Text2Live~\cite{text2live} and StableVideo~\cite{stablevideo} incorporate text-based image editing into the LNA~\cite{lna} framework. 
In another direction, Deformable Sprites~\cite{deformable} estimates sprite images and their warping parameters to reconstruct an input video.

While atlas-based editing methods enable temporally-consistent video editing, their capabilities are often limited to simple videos featuring a single foreground object with mild movement.
It is primarily due to the unreliable foreground separation process performed using an off-the-shelf segmentation model~\cite{maskrcnn,samtrack}, and atlas optimization for all moving foreground objects.
Meanwhile, RNA eliminates the need for foreground separation and employs a single atlas where the editing will occur, thereby achieving robust video editing that is applicable to a broader range of videos.

\paragraph{Propagation-based Video Editing}
Video propagation refers to general techniques that address video-related problems in a temporally-consistent manner.
It involves selecting a key frame from a video, applying a specific action to that frame, and then propagating it to the remaining video frames.
For example, Jampani et al.~\cite{videopropa} and Oh et al.~\cite{video-object-seg} propose video object mask segmentation methods, and Wang et al.~\cite{label-propa} introduce a more general label propagation method that covers object masks, textures, and human poses.
For video editing, Meyer et al.~\cite{phase-based} propose a phase-based modification transfer, and Texler et al.~\cite{video-style} introduce an editing method using patch-based training with a few shots.
These methods often show limited editing capabilities, since they largely rely on a single frame rather than exploiting multiple frames.
For example, Meyer et al.'s method is applicable only to static videos, and Texler et al.'s is limited to video stylization.



\section{ROI-based Neural Atlas} 
\label{sec:method}


In this section, we first present our ROI-based neural atlas model, which serves as the foundation for our video editing framework (\cref{sec:rna}).
Then, we explain the three main components of our video editing framework: atlas estimation (\cref{sec:atlas_estimation}), mask refinement (\cref{sec:mask_refinement}), and video reconstruction using a soft neural atlas model (\cref{sec:natural_blending}).

\subsection{ROI-based Neural Atlas Model}
\label{sec:rna}

For temporally-consistent editing of a local region in a video,
our method takes an input video, and an ROI specified by a user in a reference frame as input.
Then, our method estimates a 2D atlas representing the temporally-invariant appearance of the ROI, and the mapping from each video frame to the atlas.
Then, a user edits the 2D atlas using an image editing software such as Adobe Photoshop.
Finally, an edited video is reconstructed from the edited atlas and the original video input.
We assume that the ROI can be of any arbitrary shape, and that the ROI in the reference frame has no occluded pixels, while they can still be occluded by other objects in other frames.

To this end, we propose an ROI-based neural atlas model defined as:
\begin{equation}
  \begin{aligned}
    \hat{c}_p = \mathbb{M}(p) \mathbb{L}(\mathbb{T}(p),t) \mathbb{A}(\mathbb{T}(p)) + (1 - \mathbb{M}(p))\,c_p,
  \end{aligned}
  \label{eq:roi_model_1}
\end{equation}
where $p=(x,y,t)$ is a coordinate indicating the spatial position $(x,y)$ at the $t$-th frame.
$c$ and $\hat{c}$ are the input and its reconstructed video, respectively.
$c_p$ and $\hat{c}_p$ are the pixel values of $c$ and $\hat{c}$ at $p$, respectively.
$\mathbb{M}(p)$ is a mask indicating whether the pixel $(x,y)$ at the $t$-th frame belongs to the ROI, i.e., $\mathbb{M}(p)=1$ if $p$ belongs to the ROI, and $\mathbb{M}(p)=0$ otherwise.
$\mathbb{A}$ is an atlas representing the color values of the ROI. 
$\mathbb{A}$ is defined as a mapping from a 2D coordinate $(u,v)$ to an RGB value.
$\mathbb{T}$ is a mapping from $p$ to a coordinate $(u,v)$ on the atlas.
$\mathbb{L}$ is a scaling function to model the spatial and temporal illumination change~\cite{hashingnvd}. Specifically, $\mathbb{L}(\mathbb{T}(p),t)$ is a $3\times3$ diagonal matrix whose diagonal entries consist of scaling factors for the RGB color channels.

\begin{figure*}[t]    
   \centering
   \includegraphics[width=0.90\textwidth]{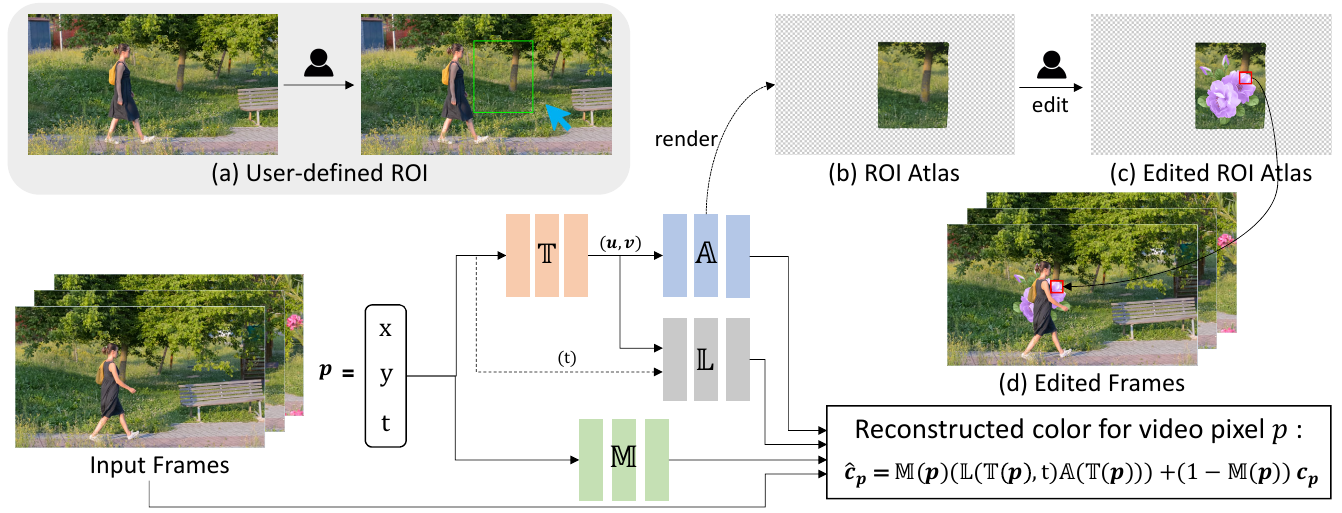}

   \vspace{-2mm}
   \caption{
   Overall framework of RNA. For video editing, (a) a user selects a reference frame from an input video and specifies an ROI where they want to edit. (b) For the specified ROI, our method estimates a 2D atlas representing its temporally-invariant appearance. (c) Then, the user edits the 2D atlas. (d) Finally, an edited video is reconstructed from the edited atlas and the input video.
   }
   \vspace{-6mm}
   \label{fig:overview}
\end{figure*}

\cref{fig:overview} illustrates our ROI-based neural atlas model in \cref{eq:roi_model_1}.
The ROI can be occluded by other objects that were originally outside of the ROI at the reference frame, but moved into the ROI later, e.g., a person passing in front of the ROI specified on the background as shown in \cref{fig:overview}.
$\mathbb{M}$ allows for handling such occlusions without explicitly modeling occluding objects. 
For the mappings $\mathbb{M}$, $\mathbb{A}$, $\mathbb{T}$, and $\mathbb{L}$, we adopt multi-layer perceptrons (MLPs). Also, we use hash encoding~\cite{instantngp} for $\mathbb{M}$ and $\mathbb{A}$ following the recent neural atlas-based approaches~\cite{hashingnvd,codef}.

To edit a video using the model in \cref{eq:roi_model_1}, we first estimate the mappings $\mathbb{M}$, $\mathbb{A}$, $\mathbb{T}$, and $\mathbb{L}$ for a given video in an end-to-end supervised manner (\cref{sec:atlas_estimation}).
Then, we perform an additional mask refinement process to more accurately consider occlusions caused by foreground objects in motion (\cref{sec:mask_refinement}).
Following this, we render a discretized version of the atlas $\mathbb{A}$.
Users then edit this atlas as they desire, and obtain an edited atlas image $\mathbb{A}_{edit}$.
Finally, an edited video is reconstructed using the video reconstruction method based on a novel soft neural atlas model, which will be described in \cref{sec:natural_blending}.

\subsection{Atlas Estimation}
\label{sec:atlas_estimation}

To edit a video, all the mappings $\mathbb{M}$, $\mathbb{A}$, $\mathbb{T}$, and $\mathbb{L}$ in Eq.~\eqref{eq:roi_model_1} are estimated in an end-to-end self-supervised manner prior to editing.
The estimation is performed by minimizing a loss $\mathcal{L}$, which is defined as:
\begin{equation}
\label{eq:total}
    \mathcal{L}=\mathcal{L}_{recon}+\mathcal{L}_{rigid} + 
    \mathcal{L}_{pos} + \mathcal{L}_{corr} + \mathcal{L}_{mask} + \mathcal{L}_{illum}
\end{equation}
where $\mathcal{L}_{recon}$, $\mathcal{L}_{rigid}$, $\mathcal{L}_{pos}$, $\mathcal{L}_{corr}$, $\mathcal{L}_{mask}$, and $\mathcal{L}_{illum}$ represent the reconstruction, rigidity, position, correspondence, mask, and illumination losses, respectively.
In the following, each loss will be elaborated in detail.

\paragraph{Reconstruction loss}
The reconstruction loss $\mathcal{L}_{recon}$ is used for estimating the mappings that can accurately reconstruct the input video.
$\mathcal{L}_{recon}$ is defined as:
\begin{equation}
\label{eq:recon}
  \begin{aligned}
   \mathcal{L}_{recon} = \sum_{p \in \mathcal{P}} \|\hat{c}_p - c_p\|^2_2 ,
  \end{aligned}
\end{equation}
where $\mathcal{P}$ is a set of pixels $p$ in an input video, including those both inside and outside the ROI. During the estimation process, we randomly sample pixels for $\mathcal{P}$ in every epoch.

\paragraph{Rigidity loss}
Estimating the mappings only with the reconstruction loss may result in a severely distorted atlas, making editing challenging.
To tackle this, we adopt the rigidity loss proposed by Kasten et al.~\cite{lna}, which is defined as:
\begin{equation}
  \begin{aligned}
    \mathcal{L}_{rigid} = \lambda_{rigid} \sum_{p\in \mathcal{P}}(
    \|J^T_{p} J_{p}\|_F +
    \|(J^T_{p} J_{p})^{-1}\|_F ),
  \end{aligned}
\end{equation}
where $\|\cdot\|_F$ denotes the Frobenius norm, $J_{p}$ is the Jacobian matrix of the local transformation at $p$ obtained from $\mathbb{T}$, and $\lambda_{rigid}$ is a balancing weight for $\mathcal{L}_{rigid}$.
The rigidity loss encourages local transformations of $\mathbb{T}$ to be as rigid as possible by enforcing the singular values of the Jacobians to be close to 1. We refer the readers to the Supplemental Document for more details of $J_p$.


\begin{figure}[t]    
   \centering
   \includegraphics[width=0.8\linewidth]{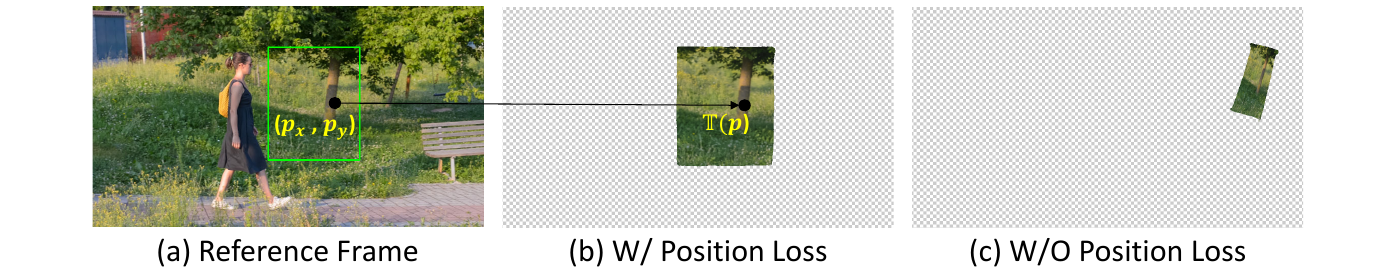}
   \vspace{-2mm}
   \caption{
   (b) Atlas estimation with $\mathcal{L}_{pos}$ provides a user-friendly interface for editing, (c) while, without $\mathcal{L}_{pos}$, the estimated 2D atlas can be severely distorted, leading to less intuitive editing.
   }
   \vspace{-1mm}
   \label{fig:position_loss}
\end{figure}

\paragraph{Position loss}
While the rigidity loss can prevent severe local distortions, resulting atlases may still be smoothly distorted. Moreover, it does not prevent global scaling or rotation of the atlas, and may result in a too small and rotated atlas, which is difficult to edit, as shown in \cref{fig:position_loss} (c).
Thus, to support more user-friendly and convenient editing, we introduce a position loss $\mathcal{L}_{pos}$, which promotes the positions of the content in the atlas to be similar to those of the user-specified ROI in the reference video frame, so that it can prevent global scaling and rotation as shown in \cref{fig:position_loss} (b).
$\mathcal{L}_{pos}$ is defined as:
\begin{equation}
  \begin{aligned}
    \mathcal{L}_{pos} = \lambda_{pos} \sum_{p\in \mathcal{P}_{ROI} }\|\mathbb{T}(p) - [x,y]^T\|_2,
  \end{aligned}
\end{equation} 
where $\lambda_{pos}$ is a balancing weight for $\mathcal{L}_{pos}$,
$\mathcal{P}_{ROI}$ is the set of pixels in the ROI in the reference frame,
and $p=(x,y,t)$.


\paragraph{Correspondence loss}
The correspondence loss $\mathcal{L}_{corr}$ ensures that the corresponding pixels across different video frames in the ROI are mapped to the same UV coordinate in the atlas.
To this end, $\mathcal{L}_{corr}$ is defined as:
\begin{equation}
  \begin{aligned}
     \mathcal{L}_{corr} =
      \lambda^{(1)}_{corr}\!\!\!\!\!\!\sum_{(p,p_r)\in\mathcal{P}_r}\!\!\!\!\!\!\mathbb{M}(p) \| \mathbb{T}(p) - \mathbb{T}(p_{r}) \|_2
     + \lambda^{(2)}_{corr}\!\!\!\!\!\!\sum_{(p,p_a)\in\mathcal{P}_a}\!\!\!\!\!\!\mathbb{M}(p) \| \mathbb{T}(p) - \mathbb{T}(p_{a}) \|_2,     
  \end{aligned}
  \label{eq:L_corr}
\end{equation}
where $\lambda^{(1)}_{corr}$ and $\lambda^{(2)}_{corr}$ are balancing weights.
$\mathcal{P}_r$ and $\mathcal{P}_a$ are sets of corresponding pixel pairs.
Specifically, $\mathcal{P}_r=\{(p,p_r)\}$ is a set such that $p$ is a pixel in the input video and $p_r$ is its match in the reference frame.
Similarly, $\mathcal{P}_a=\{(p,p_a)\}$ is a set such that $p$ is a pixel in the input video and $p_a$ is its match in an adjacent frame.
$\mathcal{P}_r$ and $\mathcal{P}_a$ can be estimated using an off-the-shelf optical flow model~\cite{raft}. 
For more accurate estimation of correspondences between distant frames, we aggregate multiple optical flow estimations. Further details can be found in the Supplemental Document.
The first term on the right hand side in \cref{eq:L_corr} ensures that the UV coordinate of a point in any given frame matches the UV coordinate of its corresponding point at the reference frame, while the second term is adopted for improving the temporal consistency of the UV coordinates between adjacent frames.

\paragraph{Mask loss}
The goal of the mask loss $\mathcal{L}_{mask}$ is to train the mask network $\mathbb{M}$ to output 1 for non-occluded pixels within the ROI and 0 otherwise.
We define $\mathcal{L}_{mask}$ as:
\begin{eqnarray}
\label{eq:mask}
    \mathcal{L}_{mask} &=& 
        - \lambda^{(1)}_{mask}\left\{\sum_{p\in\mathcal{P}_{ROI}} \log \mathbb{M}(p)
     +\sum_{p\in\mathcal{P}_{ROI}^c} \log\left(1-\mathbb{M}(p)\right)\right\} \nonumber\\
     &+&\lambda^{(2)}_{mask}\!\!\!\!\!\! \sum_{(p,p_r)\in \mathcal{P}_r}\!\!\!\!\!\! \left|\mathbb{M}(p)-\mathbb{M}(p_{r})\right| 
    +\lambda^{(3)}_{mask}\!\!\!\!\!\! \sum_{(p,p_a)\in \mathcal{P}_a}\!\!\!\!\!\! \left|\mathbb{M}(p)-\mathbb{M}(p_{a})\right|,
\end{eqnarray}
where $\lambda_{mask}^{(1)}$, $\lambda_{mask}^{(2)}$, and $\lambda_{mask}^{(3)}$ are balancing weights.
In the first term, $\mathcal{P}_{ROI}^c$ is a set of the pixels outside the ROI in the reference frame.
The first term on the right hand side is a binary cross-entropy loss to train $M$ to have 1 if the pixel is inside the ROI and 0 otherwise.
The second and third terms propagate the mask values to different frames in the input video based on the correspondence $\mathcal{P}_r$ and $\mathcal{P}_a$. 
As we assume that the ROI in the reference frame has no occluded pixels, we enforce $M$ to be 1 for all the pixels inside the ROI in the reference frame using the first term.
Then, $M$ is trained to be $0$ for the occluded pixels inside the ROI in different frames by the reconstruction loss and the second and third terms of the mask loss.

\paragraph{Illumination loss}
The illumination loss $\mathcal{L}_{illum}$ ensures that the mapping $\mathbb{L}$ accurately models changes in lighting over time.
For $\mathcal{L}_{illum}$, we adopt the residual regularization loss of Chan et al.~\cite{hashingnvd}.
Specifically, assuming that the illumination does not change much from that of the reference frame, we define $\mathcal{L}_{illum}$ as:
\begin{equation}
  \begin{aligned}
    \mathcal{L}_{illum} = \lambda_{illum} \sum_{p\in \mathcal{P}}\|\mathbb{L}(\mathbb{T}(p),t) - [1,1,1]^T\|^2_2,
  \end{aligned}
\end{equation}
where $\lambda_{illum}$ is a balancing weight.

\subsection{Mask Refinement}
\label{sec:mask_refinement}



\begin{figure}[t]    
   \centering
   \includegraphics[width=0.99\linewidth]{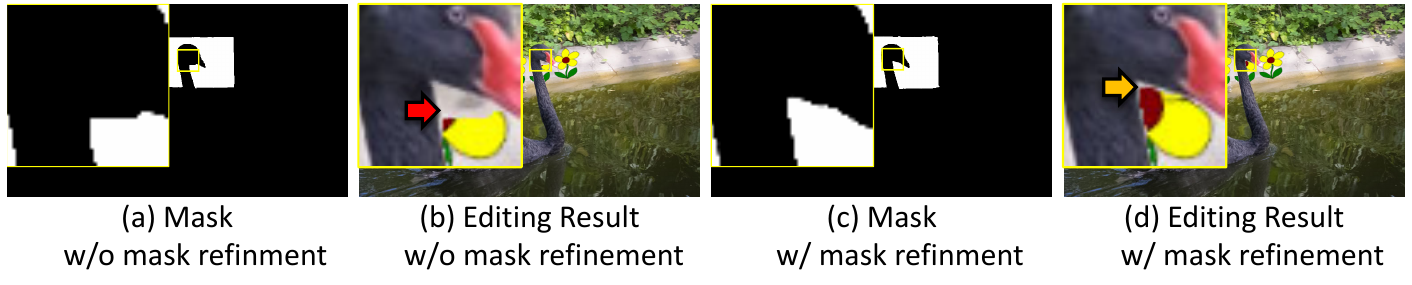}
   \vspace{-4mm}
   \caption{ 
   Magnified masks and editing results with and without mask refinement.
   Without the mask refinement, the mask inaccuracy is significant at the boundaries of occluding object.
   }   
   \vspace{-4mm}
   \label{fig:mask_refine}
\end{figure}

\begin{wrapfigure}[7]{R}{0.5\textwidth}
    \centering
    \vspace{-15mm}
    \includegraphics[width=0.5\textwidth]{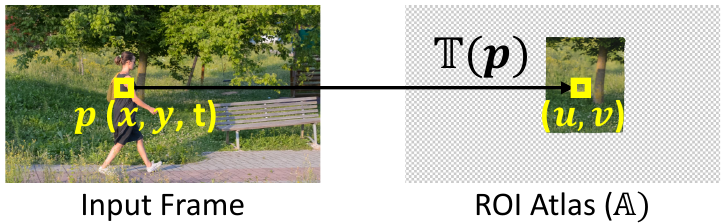} 
    \caption{A point $p$, which is outside of the ROI, is mapped into a plausible position within the ROI by $\mathbb{T}$.}
    \label{fig:mask_refine_small}
\end{wrapfigure}

While our atlas estimation process described in \cref{sec:atlas_estimation} accurately estimates $\mathbb{T}$, $\mathbb{A}$, $\mathbb{M}$, and $\mathbb{L}$ for most pixels, $\mathbb{M}$ can still contain errors, particularly around the boundaries of occluding objects, as shown by the red arrow in \cref{fig:mask_refine}(b). 
This artifact occurs when non-occluded pixels are erroneously identified as occluded ones due to the difficulty in reliably estimating a mask for these pixels.
To handle such boundary errors and to obtain a more accurate mask, we propose a novel mask refinement method, which is performed after atlas estimation. 
To this end, we leverage well-estimated $\mathbb{A}$ and $\mathbb{T}$, as shown in the example in \cref{fig:mask_refine_small}.
In the example, the point $p$, which was originally inside the ROI at the reference frame, is occluded by a foreground object.
Nevertheless, $\mathbb{T}$ still learns to map $p$ to a position inside the ROI due to the non-occluded counterparts of $p$ at other frames.
This is because $\mathbb{T}$ is modeled as an MLP, which is a piece-wise continuous function.
Based on this property, we perform an additional occlusion test. Specifically, we test whether $c_p$ is close enough to $\mathbb{A}(\mathbb{T}(p))$ to detect occluded pixels, and update the mask $\mathbb{M}$ accordingly.

\begin{figure}[t]    
   \centering
   \includegraphics[width=0.99\linewidth]{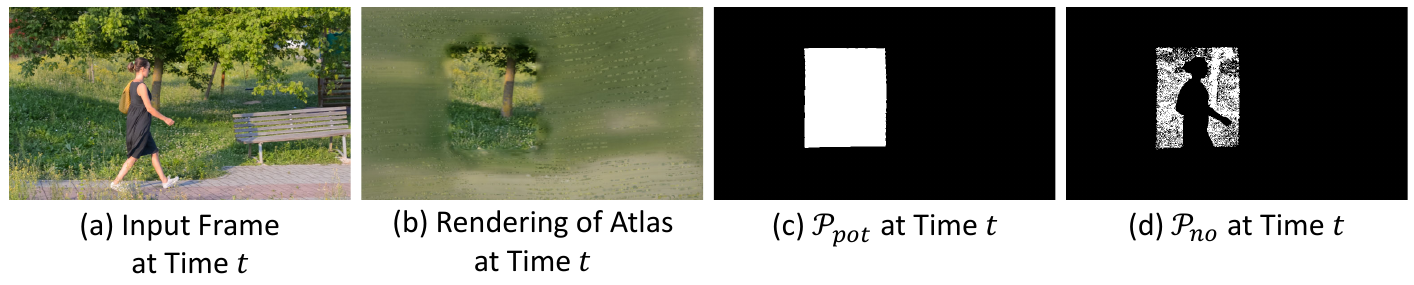}
   \vspace{-2mm}
   \caption{
   (b), which is rendered by $\mathbb{L}(\mathbb{T}(p),t) \mathbb{A}(\mathbb{T}(p))$, shows an appearance without the moving foreground object that exists in the input frame at $t$. (d) Using this property, we can effectively identify the target points, $\mathcal{P}_{no}$, for mask refinement.
   }   
   \vspace{-5mm}
   \label{fig:refine_points}
\end{figure}

Specifically, in the first stage of the refinement step,
we first find a set of pixels in the input video potentially belonging to the ROI.
To this end, we build a binary mask $M_{roi}$ for the reference frame such that $M_{roi}(x,y) = 1$ if $(x,y)$ is inside the ROI at the reference frame, and $M_{roi}(x,y)=0$ otherwise.
Then, we warp $M_{roi}$ using the transformation $\mathbb{T}$ for the reference frame, and obtain a mask $M_{roi}^{w}$ that is aligned to the atlas $\mathbb{A}$.
Using $M_{roi}^w$, we find a set of pixels potentially belonging to the ROI as $\mathcal{P}_{pot} = \left\{ p | M_{roi}^w(\mathbb{T}(p)) = 1 \right\}$ (\cref{fig:refine_points} (c)).
We then identify non-occluded pixels among the pixels in $\mathcal{P}_{pot}$.
Specifically, we find a set of non-occluded pixels $\mathcal{P}_{no}$
as $\mathcal{P}_{no}= \left\{ p | \|\mathbb{L}(\mathbb{T}(p),t) \mathbb{A}(\mathbb{T}(p))-c_p \|_1 < \tau \right\} \cap \mathcal{P}_{pot}$
where $\tau$ is a small constant.
\cref{fig:refine_points} (d) visualizes an example of $\mathcal{P}_{no}$.

The next stage of the refinement step updates the mask $\mathbb{M}$ by minimizing the loss function defined as:
\begin{equation}
    \begin{aligned}
        \mathcal{L}_{refine}=\mathcal{L}_{recon} + \mathcal{L}_{no},
    \end{aligned}
\end{equation}
where $\mathcal{L}_{no}$ is a loss for the non-occluded pixels.
$\mathcal{L}_{no}$ is defined as:
\begin{equation}
    \begin{aligned}
        \mathcal{L}_{no} = 
	  \lambda_{no}^{(1)} \sum_{p\in \mathcal{P}_{no}} |1-\mathbb{M}(p)|^2 
        +\lambda_{no}^{(2)} \sum_{(p,p_a)\in \mathcal{P}_a}\!\!\!\! \left| \mathbb{M}(p) - \mathbb{M}(p_a) \right|,
    \end{aligned}
\end{equation}
where $\lambda^{(1)}_{no}$ and $\lambda^{(2)}_{no}$ are balancing weights.
The first term on the right hand side promotes $\mathbb{M}$ to be close to 1 for the non-occluded pixels inside the ROI while the second term propagates the refined mask values to other frames.
We update only $\mathbb{M}$ using $\mathcal{L}_{refine}$ in the mask refinement step.

\subsection{Video Reconstruction using a Soft Neural Atlas Model}
\label{sec:natural_blending}


Although our mask refinement step can effectively enhance the accuracy of the mask $\mathbb{M}$,
the mask after the refinement tends to be close to a hard mask whose values are either 0 or 1.
On the other hand, pixels along the boundaries between occluding objects and the ROI usually have mixtures of the colors from different regions.
Thus, using a hard mask leads to unnatural reconstruction results after atlas editing as indicated by the orange arrow in \cref{fig:mask_refine} (d).

For more visually pleasing blending, we thus estimate a soft mask that correctly reflects the blending of occluding objects and the ROI using an off-the-shelf matting network, VitMatte \cite{vitmatte}.
Specifically, for each video frame, we render its mask from $\mathbb{M}$, and apply morphological erosion and dilation to obtain a trimap. Then, we estimate a soft mask $\hat{\mathbb{M}}$ by feeding the frame and its trimap to the matting network.

For video reconstruction after atlas editing, we may na\"ively replace $\mathbb{A}$ and $\mathbb{M}$ in \cref{eq:roi_model_1} with $\mathbb{A}_{edit}$ and $\hat{\mathbb{M}}$, respectively.
However, this approach does not produce natural-looking results on the boundary pixels around occluding objects since \cref{eq:roi_model_1} is designed in the consideration of using a hard mask.
%
To address this, we additionally propose a soft neural atlas model, which is defined as:
\begin{equation}
    c_p = \hat{\mathbb{M}}(p)\mathbb{L}(\mathbb{T}(p),t)\mathbb{A}(\mathbb{T}(p))+(1-\hat{\mathbb{M}}(p))c_p^{oc},
    \label{eq:soft_roi_model}
\end{equation}
where $c_p^{oc}$ is the color of the occluding object at $p$.
From \cref{eq:soft_roi_model}, we can derive $c_p^{oc}$ as:
\begin{equation}
    c_p^{oc} = \frac{c_p - \hat{\mathbb{M}}(p)\mathbb{L}(\mathbb{T}(p),t)\mathbb{A}(\mathbb{T}(p))}{1-\hat{\mathbb{M}}(p)}.
    \label{eq:c_p_oc}
\end{equation}
Then, by replacing $\mathbb{A}$ in \cref{eq:soft_roi_model} and substituting \cref{eq:c_p_oc} into \cref{eq:soft_roi_model},
we obtain our final reconstruction equation, which is defined as:
\begin{equation}
    \label{eq:blending2}
    \begin{aligned}
        c_p^{edit} = 
        \begin{cases}
         \hat{\mathbb{M}}(p)\mathbb{L}(\mathbb{T}(p),t)(\mathbb{A}_{edit}(\mathbb{T}(p))-\mathbb{A}(\mathbb{T}(p)))+c_p  &  \hat{\mathbb{M}}(p) < 1 \\
         \mathbb{L}(\mathbb{T}(p),t)\mathbb{A}_{edit}(\mathbb{T}(p))     &  \hat{\mathbb{M}}(p) = 1, \\
        \end{cases}
    \end{aligned}
\end{equation}
where $c^{edit}$ is a video reconstructed from $\mathbb{A}_{edit}$.
The equation for $\hat{\mathbb{M}}(p)<1$ in \cref{eq:blending2} should reduce to $\mathbb{L}(\mathbb{T}(p),t)\mathbb{A}_{edit}(\mathbb{T}(p))$ when $\hat{\mathbb{M}}(p) = 1$ if $\mathbb{A}$ and $\mathbb{L}$ are perfectly estimated, i.e., $\mathbb{L}(\mathbb{T}(p),t)\mathbb{A}(\mathbb{T}(p)) = c_p$.
However, $\mathbb{A}$ and $\mathbb{L}$ may have a small amount of error in practice, so we use $\mathbb{L}(\mathbb{T}(p),t)\mathbb{A}_{edit}(\mathbb{T}(p))$ for $\hat{\mathbb{M}}(p) = 1$.

\begin{figure}[t]    
   \centering
   \includegraphics[width=0.99\textwidth]{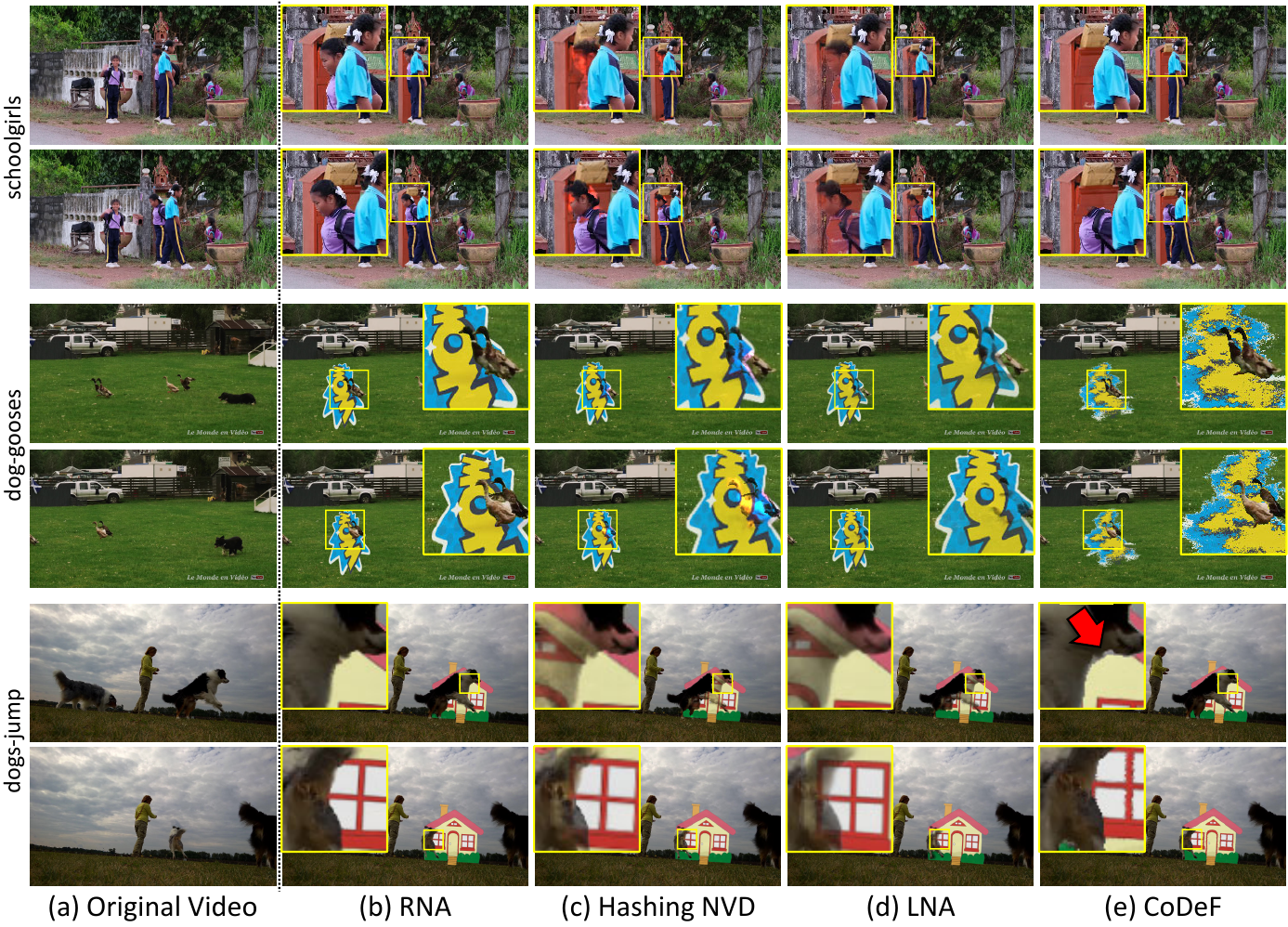}
   \vspace{-2mm}
   \caption{Qualitative comparisons with previous methods. RNA achieves natural editing by handling complex occlusions without artifacts, such as ghosting and omission, significantly outperforming previous methods.
   }
   \label{fig:cmp1}
   \vspace{-1mm}
\end{figure}

\begin{figure*}[t]    
   \centering
   \includegraphics[width=0.99\textwidth]{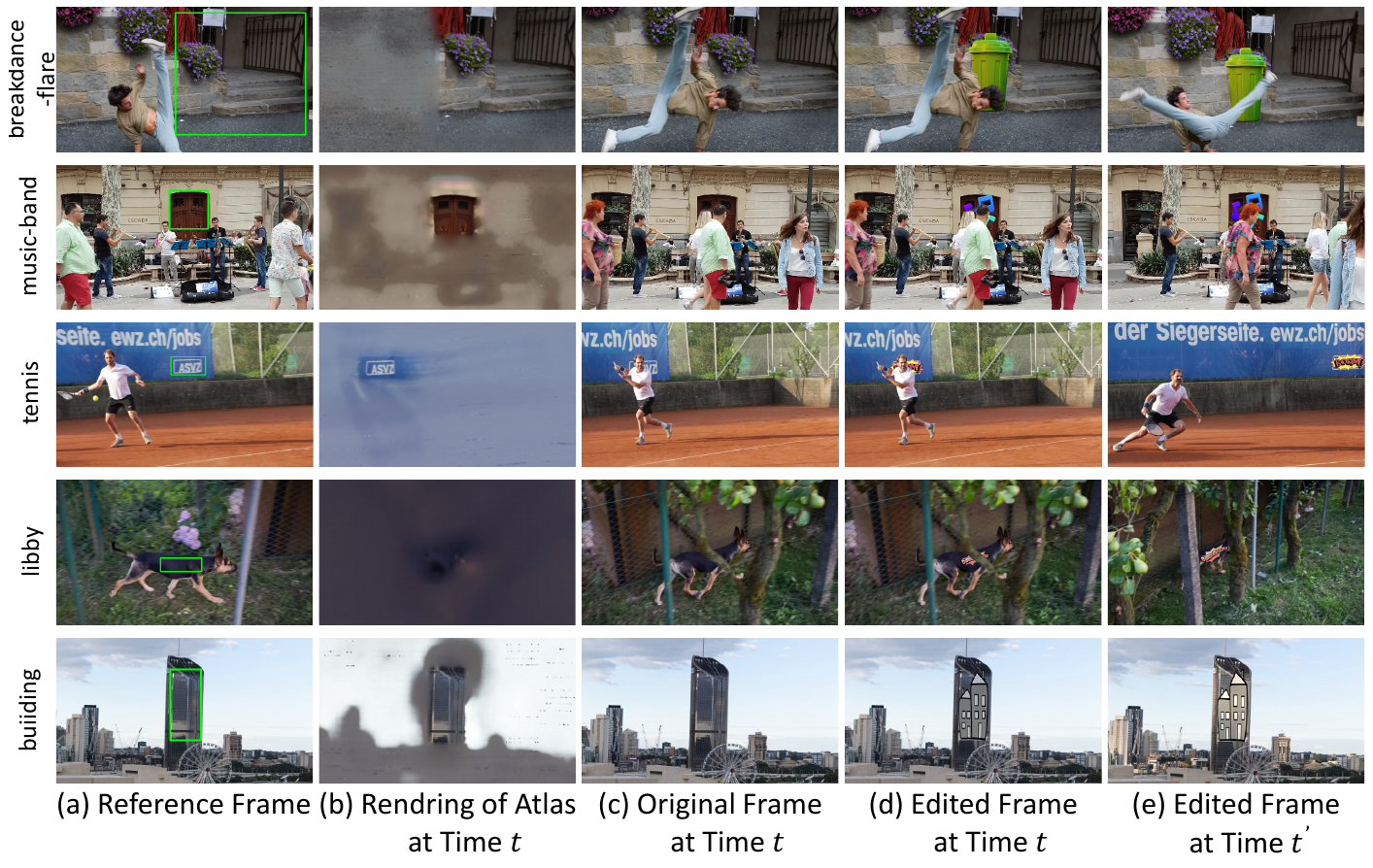}
   \vspace{-1mm}
   \caption{
   Editing results of RNA applied to various video scenarios, including complex foreground objects (breakdance-flare, music-band), large camera movements (tennis), foreground object with occlusion (libby) and time-varying light change (building).
   }
   \vspace{-1mm}
   \label{fig:qual1}
\end{figure*}

\section{Experiments}
\label{sec:exp}

In this section, we conduct extensive experiments to demonstrate the effectiveness of RNA.
We use video examples from the DAVIS dataset \cite{davis} for the evaluations.
We refer the readers to the Supplemental Document for implementation details including the network architecture of the mappings, the balancing weights of the loss function, and the details of the editing process.

\begin{table}[t]
\centering

\caption{
Quantitative comparisons of memory usage, training time, and PSNR for reconstructed frame. 
The training is conducted using a GeForce RTX 3090 GPU with 24 GB.
Deformable Sprites~\cite{deformable} is tested at a lower resolution due to memory constraints.
}

\scalebox{0.65}{
\begin{tabular}{c|c|ccc|ccc|ccc}
\Xhline{2\arrayrulewidth}

\multicolumn{1}{l}{} \vline& \multicolumn{1}{l}{} \vline& \multicolumn{3}{c}{lucia [1 object, 70 frames]}               \vline& \multicolumn{3}{c}{dogs-jump [3 objects, 65 frames]}                                                            \vline& \multicolumn{3}{c}{dog-gooses [5 objects, 70 frames]}                                                                                                  \\
Method               & Resolution           &

\begin{tabular}[c]{@{}c@{}}GPU   \\      Memory\end{tabular} & \begin{tabular}[c]{@{}c@{}}Training   \\      Time\end{tabular} & PSNR & \begin{tabular}[c]{@{}c@{}}GPU   \\      Memory\end{tabular} & \begin{tabular}[c]{@{}c@{}}Training   \\      Time\end{tabular} & PSNR & \begin{tabular}[c]{@{}c@{}}GPU   \\      Memory\end{tabular} & \begin{tabular}[c]{@{}c@{}}Training   \\      Time\end{tabular} & PSNR \\\hline\hline
Deformable   Sprites~\cite{deformable}  & 427 $\times$  240            & 9.4 GB                                                       & 25 min.                                                      & 25.6 & 15.1 GB                                                       & 35 min.                                                      & 30.5 & 21.2 GB                                                       & 55 min.                                                      & 25.3 \\
CoDeF~\cite{codef}                & 768 $\times$ 432            & 3.8 GB                                                        & 10 min.                                                      & 26.2 & 3.8 GB                                                        & 10 min.                                                      & 33.8 & 3.8 GB                                                        & 10 min.                                                      & 25.1 \\
LNA~\cite{lna}                  & 768 $\times$ 432            & 3.1 GB                                                        & 6 hours                                                         & 31.4 & 4.7 GB                                                        & 10 hours                                                        & 33.8 & 6.3 GB                                                        & 14 hours                                                        & 27.6 \\
Hashing NVD~\cite{hashingnvd}          & 768 $\times$ 432            & 2.9 GB                                                        & 1.2 hours                                                       & 31.0   & 3.7 GB                                                        & 2.2 hours                                                       & 33.2 & 4.4 GB                                                        & 3.2 hours                                                       & 27.3 \\\hline
Ours                 & 768 $\times$ 432            & 2.4 GB                                                        & 40 min.                                                      & 31.5 & 2.4 GB                                                        & 40 min.                                                      & 33.5 & 2.4 GB                                                        & 40 min.                                                      & 27.7 \\
\Xhline{2\arrayrulewidth}

\end{tabular}
}

\vspace{-1mm}
\label{tbl:recon}
\end{table}

\subsection{Video Editing Quality}
\label{sec:cmp-edit}

We first compare the quality of video editing results of different methods.
\cref{fig:cmp1} shows a qualitative comparison of the editing results of Hashing NVD \cite{hashingnvd}, LNA \cite{lna} and CoDeF \cite{codef}.
Hashing NVD \cite{hashingnvd} and LNA \cite{lna} exhibit ghosting artifacts in all editing results due to the inaccurate atlas estimation of the foreground objects.
CoDeF~\cite{codef} exhibits several issues across the editing examples.
In the `schoolgirls' example, the head of the girl in violet disappears in the edited frames, because CoDeF relies on instance segmentation to extract foreground objects, which is unfortunately unreliable for objects with complex motions.
In the `dog-gooses' example, the edited contents show jittering artifacts as the camera moves due to its relatively na\"ive motion estimation. 
In the `dogs-jump' example, boundary artifacts can be observed between the dog and the background due to its inaccurate segmentation and hard mask-based approach.
In contrast, RNA achieves natural video editing results in these challenging scenarios.

\subsection{Reconstruction Quality and Efficiency}
\label{sec:cmp-recon}

For plausible video editing, accurate atlas estimation is crucial.
To evaluate the quality of atlas estimation of RNA, we compare video reconstruction results of different methods on video examples with various numbers of moving objects in \cref{tbl:recon}.
Additionally, we also compare the computation times to evaluate the efficiency of the proposed method.
In the table, we compare the GPU memory usage and the training times needed for atlas and mask estimation with those of previous methods. For a fair comparison, we
compare the PSNR value within a specified ROI region between the reference and reconstructed frames. The details of these ROI regions are included in the Supplementary Document.
As described in the table, Deformable Sprites~\cite{deformable}, LNA~\cite{lna}, and Hashing NVD~\cite{hashingnvd} require memory and training times that proportionally increase with the number of foreground objects, as they model each foreground object using an individual network. 
Meanwhile, RNA achieves comparable PSNR values to these methods with generally smaller and constant computational overload, regardless of the number of moving objects. 
CoDeF~\cite{codef} exceptionally requires constant and small computational overload, similar to our approach, since it models only a single atlas for the background region. 
However, it exhibits poor PSNR values when there are camera motions, as shown in the `lucia' and `dog-gooses' examples. This poor reconstruction results in jittering artifacts after editing, as shown in the results of `dog-gooses' and `dogs-jump' in \cref{fig:cmp1} (e).

\subsection{Additional Qualitative Examples}
\label{sec:qual}


\cref{fig:qual1} shows rendering results of estimated atlases and video editing results applied to various scenarios.
As shown in (b) and (c) in the figure, the rendering results of atlases exhibit the contents without occluding foreground objects. We effectively utilize this property in our mask refinement and soft neural atlas model, achieving high-quality editing results.
(d) and (e) in the figure demonstrate high-quality video editing results achieved by RNA. Especially, despite the high complexity with many people walking around in the `music-band' example, RNA effectively handles such challenging occlusions, achieving natural video editing results. In the `tennis' example, RNA succeeds in editing despite large camera movements.
The `libby' example shows that RNA successfully edits the moving foreground object despite its non-rigid motion, and handles occlusions despite the motion of the foreground object.
The `building' example demonstrates that RNA can also effectively handle time-varying illumination changes.

\begin{figure}[t]    
   \centering
   \includegraphics[width=0.99\linewidth]{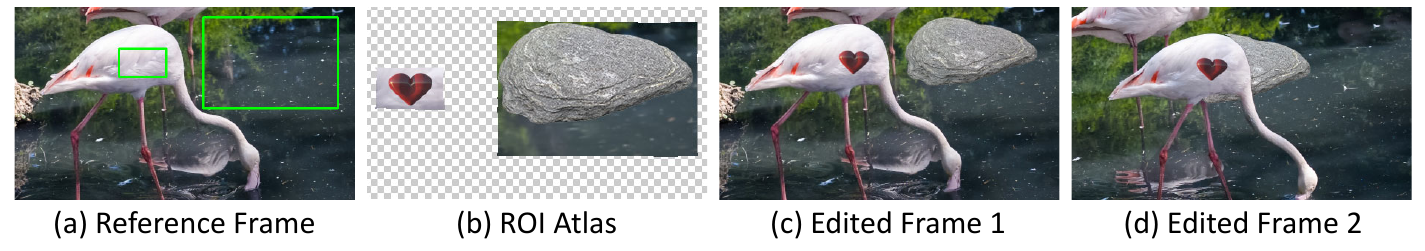}
   \vspace{-1mm}
   \caption{
   (a) shows multiple ROIs designated for a moving foreground object and a static background, and (b) shows an estimated atlas corresponding to those ROIs. 
   (c) and (d) show that RNA can address the editing in both the foreground object and the background using a single atlas. 
   }
   \vspace{-0mm}
   \label{fig:multi-roi}
\end{figure}

\paragraph{Multiple ROIs}
RNA also allows for a user to simultaneously edit both a moving foreground object and the background as long as they are not overlapped in the reference frame. For editing multiple ROIs, a user specifies multiple ROIs, as shown in \cref{fig:multi-roi}.
Then, RNA estimates a single atlas for multiple ROIs as shown in \cref{fig:multi-roi} (b) so that the user can edit the ROIs together.

\begin{figure*}[t]    
   \centering
   \includegraphics[width=1.0\linewidth]{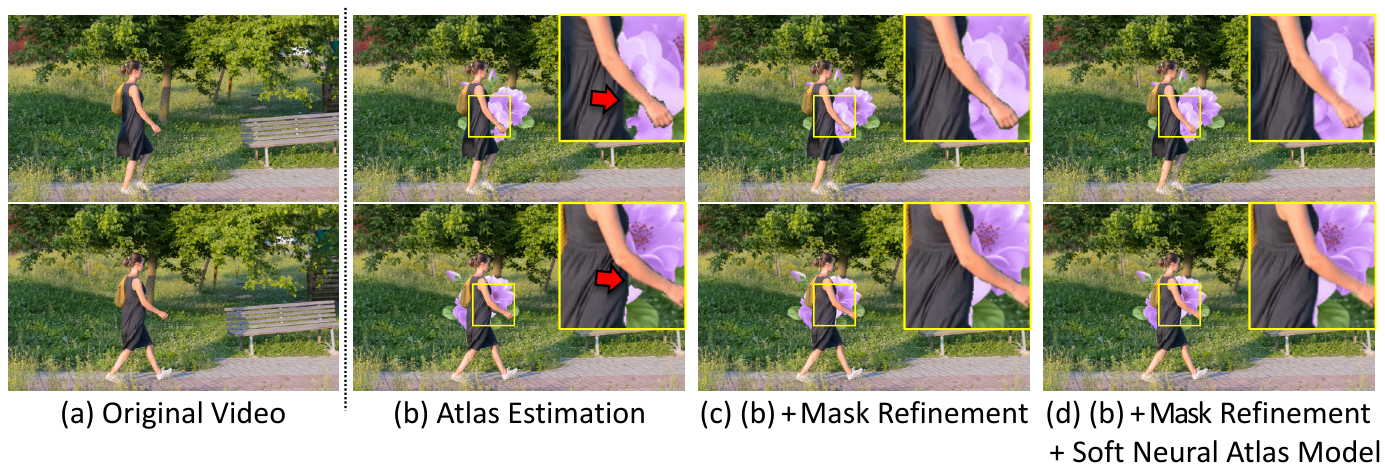}
   \vspace{-5mm}
   \caption{
   Ablation study for our altas estimation, mask refinement, and soft neural atlas model. (b) Atlas estimation produces artifacts for pixels around the boundaries of occluding objects (red arrows). (c) Mask refinement successfully addresses those artifacts. (d) Soft neural atlas model achieves smooth compositions between the editing area and the foreground. 
   }
    \vspace{-1mm}
   \label{fig:stage_ablation}
\end{figure*}

\begin{figure}[t]    
   \centering
   \includegraphics[width=1.0\linewidth]{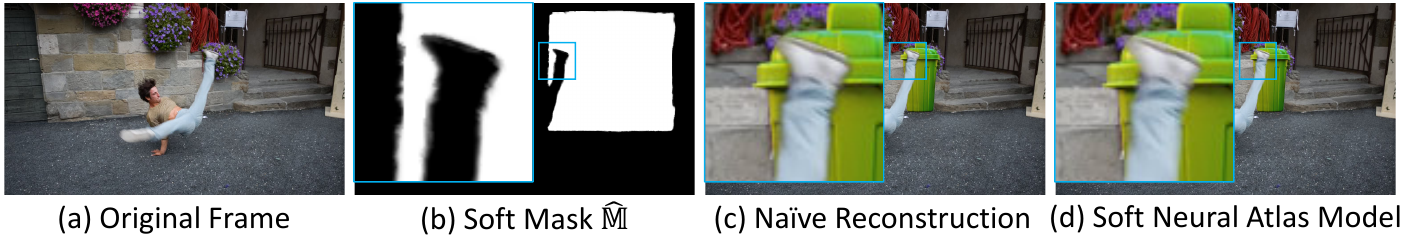}
   \vspace{-4mm}
   \caption{
   Comparison of (c) a na\"ive reconstruction result, which uses \cref{eq:roi_model_1} by replacing $\mathbb{M}$ and $\mathbb{A}$ with $\hat{\mathbb{M}}$ and $\mathbb{A}_{edit}$, and (d) a result using our soft neural atlas model.
   In the boundary areas between the edited content and the occluding foreground object, our proposed model shows a more natural-looking transition between them.
   }
   \vspace{-1mm}
   \label{fig:blending_ablation}
\end{figure}

\subsection{Ablation Study}

\paragraph{Impact of each phase}
We conduct an ablation study by sequentially applying our mask refinement and soft neural atlas model after atlas estimation (\cref{fig:stage_ablation}). Without the mask refinement phase, an imprecise mask is estimated, leading to artifacts for pixels around the boundaries of occluding objects (red arrows in \cref{fig:stage_ablation} (b)). Our mask refinement effectively suppresses these artifacts. Finally, our soft neural atlas model smoothly blends the edited content and moving foreground object, as shown in \cref{fig:stage_ablation} (d). More ablation examples are displayed to the Supplemental Document.


\paragraph{Soft neural atlas model}
We conduct another ablation study to verify the effect of our soft neural atlas model.
A na\"ive approach to video reconstruction is to use the hard mask-based neural atlas model presented in \cref{eq:roi_model_1}.
Specifically, we may simply replace $\mathbb{A}$ and $\mathbb{M}$ with $\mathbb{A}_{edit}$ and $\hat{\mathbb{M}}$, respectively, to achieve video reconstruction using an edited atlas.
In this ablation study, we compare our soft neural atlas model-based video reconstruction against the na\"ive approach to verify the effect of the soft neural atlas model in \cref{fig:blending_ablation}.
\cref{fig:blending_ablation} (c) shows a result of the na\"ive reconstruction approach.
As shown in the magnified view in \cref{fig:blending_ablation} (c), the boundary between the leg and green bin includes unnatural dark colors, which originate from the original input frame. In contrast, our soft neural atlas model successfully avoids such artifacts and achieves highly-natural reconstruction results, as shown in \cref{fig:blending_ablation} (d).

\section{Conclusions}
\label{sec:conclusion}


In this paper, we propose RNA, a novel ROI-based video editing framework. RNA enables video editing by allowing users to specify an ROI that they want to edit.
Our ROI-based approach enables computationally-efficient and robust atlas and mask estimation, while removing the burden of users to manually specify all moving foreground objects.
For high-quality video editing, we also present a novel mask refinement method, and a novel soft neural atlas model.
Consequently, RNA offers a more practical and efficient solution to video editing that is applicable to a wider range of videos.
Our method is not free from limitations. we assume that the ROI a user wants to edit must be clearly visible without any occluding objects in a reference frame. However, such frames may not be available in some videos.



\paragraph{\small {\bf Acknowledgements}}
This work was supported by Institute of Information \& communications Technology Planning \& Evaluation (IITP) grants funded by the Korea government (MSIT) (No.2019-0-01906, Artificial Intelligence Graduate School Program(POSTECH), RS-2024-00395401, Development of VFX creation and combination using generative AI).


\clearpage  

%
%
\bibliographystyle{splncs04}
\bibliography{main}

\clearpage
\pagebreak

\title{RNA: Video Editing with ROI-based Neural Atlas\\--Supplemental
Document--} 

\titlerunning{RNA}

\author{Jaekyeong Lee\thanks{equal contribution} \and
Geonung Kim\printfnsymbol{1}  \and
Sunghyun Cho} 

\authorrunning{J.~Lee et al.}

\institute{POSTECH\\
\email{\{jaekyeong,k2woong92,s.cho\}@postech.ac.kr}\\
\url{https://jaekyeongg.github.io/RNA}
}


\maketitle

\setcounter{section}{0}
\setcounter{figure}{0}
\setcounter{table}{0}

\renewcommand\thesection{S\arabic{section}}
\renewcommand\thefigure{S\arabic{figure}}
\renewcommand\thetable{S\arabic{table}}

\begin{figure*}[t]    
   \centering
   \includegraphics[width=0.99\linewidth]{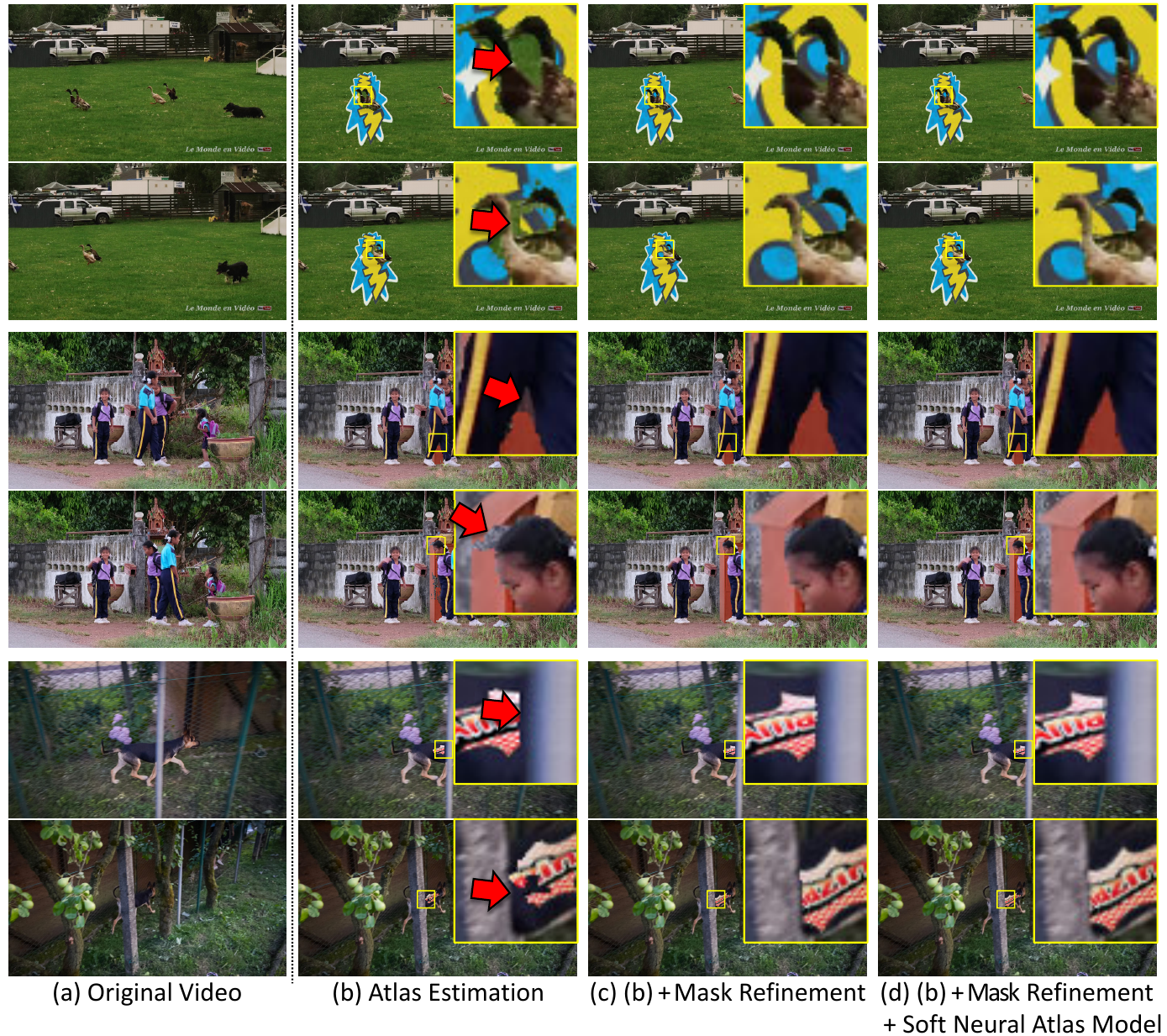}
   \vspace{-2mm}
   \caption{
   Ablation study for our altas estimation, mask refinement, and soft neural atlas model. (b) Atlas estimation produces artifacts for pixels around the boundaries of occluding objects (red arrows). (c) Mask refinement successfully addresses those artifacts. (d) Soft neural atlas model achieves smooth compositions between the editing area and the foreground. 
   }
   \vspace{-2mm}
   \label{fig:stage_ablation_2}
\end{figure*}

\begin{figure*}[t]    
   \centering
   \includegraphics[width=0.99\linewidth]{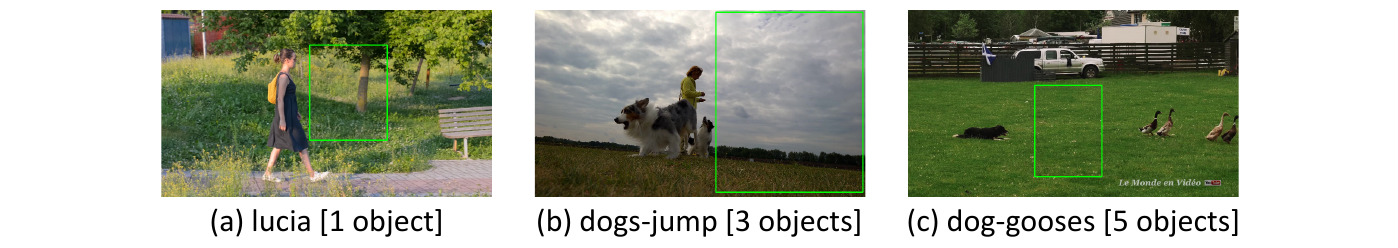}
   \vspace{-2mm}
   \caption{
   The green boxes indicate the ROI regions used for measuring PSNR in Tab.1 of the main manuscript.
   }
   \vspace{-2mm}
   \label{fig:roi_mark}
\end{figure*}

In this Supplementary Document, we present:

\begin{itemize}
  \item Multiple optical flow estimations (\cref{sec:multiple optical flow})
  \item Implementation details (\cref{sec:detail})
  \item Additional details of RNA (\cref{sec:detail2})
  \item Additional qualitative results of ablation studies (\cref{fig:stage_ablation_2})
  \item ROI regions used for measuring PSNR in the main paper
 (\cref{fig:roi_mark})
\end{itemize}

\section{Multiple Optical Flow Estimations}
\label{sec:multiple optical flow}
\begin{figure*}[t]    
   \centering
   \includegraphics[width=0.99\linewidth]{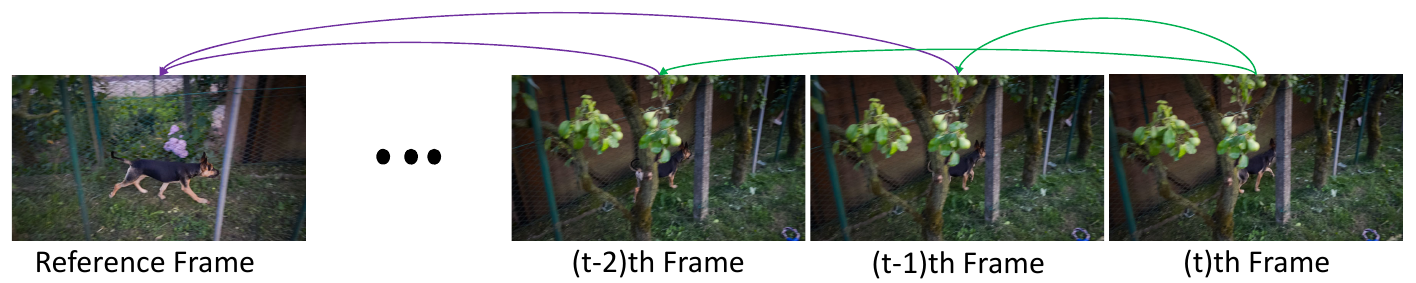}
   \vspace{-2mm}
   \caption{Multiple optical flow estimations. The green line represents the flow estimation $f_{t\rightarrow t'}$ from $t$-th frame to the temporal neighbor frame, while the purple line indicates the flow estimation $F_{t'\rightarrow r}$ from the temporal neighbor frame to the reference frame.}
   \vspace{-2mm}
   \label{fig:multiple_opticalflow}
\end{figure*}

As described in Sec~3.2 in the main paper, our framework requires the optical flow from each video frame to the reference frame. However, optical flow estimation may fail at some frames, especially at distant frames, resulting in quality degradation in the atlas estimation. To address the inaccurate na\"ive estimation, we employ multiple optical flow estimations instead of relying on a single estimation, as shown in \cref{fig:multiple_opticalflow}. Suppose that $f_{t\rightarrow t'}$ is an optical flow map from the $t$-th frame to the $t'$-th frame.
We first remove inaccurate flow vectors from $f_{t\rightarrow t'}$ by using the cycle consistency and appearance tests as done in OmniMotion~\cite{trackingeverything}. Specifically, we filter out flows with a forward-backward flow error greater than 5 pixels for cycle consistency and flows with a cosine similarity between the DINO~\cite{caron2021emerging} features less than 0.5 for the appearance test, respectively.
Then, we obtain a refined optical flow map $F_{t\rightarrow r}$ from the $t$-th frame to the reference frame as $F_{t\rightarrow r}=\textrm{avg}_{t'\in T_t}\left(f_{t\rightarrow t'}\circ F_{t'\rightarrow r}\right)$ where $T_t$ is a temporal neighborhood of the $t$-th frame, $\circ$ is a composition operator that combines two flow maps, and $\textrm{avg}$ is a pixel-wise averaging operator. 
We use a maximum total of 7 frames as the temporal neighborhood.

\section{Implementation Details}
\label{sec:detail}
For the mappings $\mathbb{M}$, $\mathbb{A}$, $\mathbb{T}$ and  $\mathbb{L}$, we adopt multi-layer perceptrons (MLPs). $\mathbb{M}$ and $\mathbb{A}$ have 4 layers with hash grid encoding, with $\mathbb{M}$ having 32 channels and $\mathbb{A}$ having 64 channels. $\mathbb{T}$ and $\mathbb{L}$ use 4 layers without hash grid encoding, comprising 256 channels and 64 channels, respectively. $\operatorname{ReLU}$ is used as an activation function between the network layers, $\operatorname{tanh}$ function is employed for the output layer of $\mathbb{M}$, $\mathbb{A}$, $\mathbb{T}$, while the $\mathbb{L}$ output layer uses the $\operatorname{softplus}$ function.
Before the atlas estimation phase, we apply a bootstrapping process for a better mask initialization, similar to LNA~\cite{lna}. 
Specifically, we obtain masks for all frames by using the optical flow from the reference frame to each of them in \cref{sec:multiple optical flow}.
We then optimize $\mathbb{M}$ using binary cross-entropy loss with the obtained masks during 10K iterations.
In the atlas estimation phase, we train all the mappings $\mathbb{M}$, $\mathbb{A}$, $\mathbb{T}$ and $\mathbb{L}$ using balancing weights of $\lambda_{rigid} = 0.002$, $\lambda_{pos} = 0.6$, $\lambda^{(1)}_{corr} = 0.02$, $\lambda^{(2)}_{corr} = 0.1$, $\lambda^{(1)}_{mask} = 0.4$, $\lambda^{(2)}_{mask} = 0.6$, $\lambda^{(3)}_{mask} = 0.4$,
$\lambda_{illum} = 0.1$ for 60K iterations.
In the mask refinement phase, we train only $\mathbb{M}$ using balancing weights of $\lambda^{(1)}_{no} = 0.01$, $\lambda^{(2)}_{no} = 0.1$ and a threshold of $\tau=0.025$ for 20K iterations.
At each iteration, we randomly sample 10K points from the input video.
We use Adam optimizer~\cite{adam} to optimize the MLPs and hash encoding with learning rate 0.0005 and 0.05, respectively. 
At a resolution of 768$\times$432 with 70 frames, using a GeForce RTX 3090 with 24 GB of memory, training requires a total of 40 minutes.
This includes 35 minutes for atlas estimation, 5 minutes for mask refinement, and 45 seconds for soft mask estimation.

\section{Additional Details of RNA}
\label{sec:detail2}

\begin{figure}[t]    
   \centering
   \includegraphics[width=0.99\linewidth]{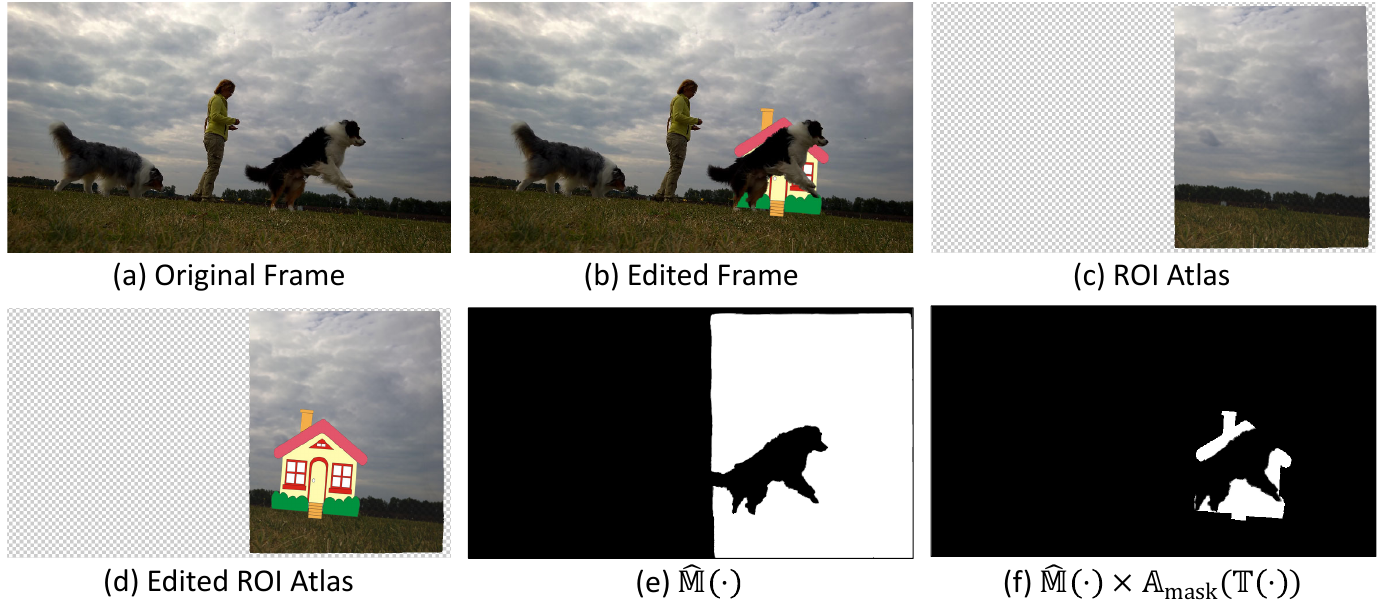}
   \vspace{-0mm}
   \caption{(f) Visualization of the advanced mask that allows more content to be retained from the original video rather than the atlas.
   }
   \vspace{-0mm}
   \label{fig:editing_mask}
\end{figure}

\paragraph{Details of atlas editing}
To edit the atlas $\mathbb{A}$, we first render the ROI atlas to a 1000$\times$1000 pixel image by feeding uv coordinates ranging from -1 to 1 into $\mathbb{A}$ (\cref{fig:editing_mask}(c)).
Users then perform edits on this image, resulting in an edited ROI Atlas $\mathbb{A}_{edit}$ (\cref{fig:editing_mask}(d)).
Unlike the continuous image $\mathbb{A}$, $\mathbb{A}_{edit}$ is a discrete image.
To treat $\mathbb{A}_{edit}$ as a continuous image, we employ bilinear interpolation when reconstructing the edited video.

\paragraph{High-quality reconstruction}
To achieve higher-quality reconstruction for the edited video, we use the original video colors for the areas outside the edited region. To this end, we first obtain an atlas mask $\mathbb{A}_{mask}$ that indicates the areas where the colors have changed in the ROI atlas due to user editing, as defined by:
\begin{equation}
    \label{eq:blending2}
    \begin{aligned}
        \mathbb{A}_{mask}(u,v) = 
        \begin{cases}
         0  &  \mathbb{A}(u,v) = \mathbb{A}_{edit}(u,v) \\
         1  &  \text{otherwise}. \\
        \end{cases}
    \end{aligned}   
\end{equation}
Then, with $\mathbb{A}_{mask}$ and $\mathbb{T}$, we can identify whether a pixel in the edited video is edited or not. 
Formally, at an arbitrary point $p$ in a video, we use $\hat{\mathbb{M}}(p) \times \mathbb{A}_{mask}(\mathbb{T}(p))$ as a final mask value for the video reconstruction. 
As shown in \cref{fig:editing_mask}(e) and (f), the mask region is effectively reduced, allowing more content to be retained from the original video rather than the atlas.
Since this advanced mask is also easily acquired to all previous methods, we applied it to all editing results, including ours and those from previous methods.


\paragraph{Jacobian matrix $J_p$ in the rigidity loss}

We approximate $J_p$ as:
\begin{equation}
    J_p=\left[\mathbb{T}(p_x)-\mathbb{T}(p)    \quad \mathbb{T}(p_y)-\mathbb{T}(p)\right] \in \mathbb{R}^{2\times2}
\end{equation}
where $p_x=(x+\delta, y, t)$ and $p_y = (x, y+\delta, t)$. $\delta$ is a small offset, which is set $1$ in our experiments.

\end{document}